\documentclass[10pt]{article}

\usepackage{booktabs}
\usepackage{multirow}
\usepackage{makecell}
\usepackage{wrapfig}
\usepackage[table]{xcolor}
\usepackage{colortbl}
\usepackage{arydshln}

\usepackage[utf8]{inputenc} 
\usepackage[T1]{fontenc}    
\usepackage{url}            
\usepackage{graphicx}       
\usepackage{amsmath}        
\usepackage{amsfonts}       
\usepackage{nicefrac}       
\usepackage{microtype}      
\usepackage{caption}
\usepackage{needspace}
\newcommand{\best}[1]{{\bfseries #1}}
\definecolor{LightGray}{gray}{0.92}
\definecolor{RefBlue}{RGB}{46,117,182}

\makeatletter
\RequirePackage{times}
\RequirePackage{graphicx}
\RequirePackage{amssymb}
\RequirePackage[numbers,square]{natbib}
\RequirePackage{authblk}
\RequirePackage[absolute,overlay]{textpos}
\RequirePackage{xcolor}
\RequirePackage{fancyhdr}
\RequirePackage{ifthen}
\RequirePackage[colorlinks=true,
  linkcolor=purple,
  urlcolor=blue,
  citecolor=cyan,
  anchorcolor=black]{hyperref}

\RequirePackage{tikz}
\definecolor{lime}{HTML}{A6CE39}
\DeclareRobustCommand{\orcidicon}{%
  \begin{tikzpicture}
  \draw[lime, fill=lime] (0,0)
  circle [radius=0.16]
  node[white] {{\fontfamily{qag}\selectfont \tiny ID}};
  \draw[white, fill=white] (-0.0625,0.095)
  circle [radius=0.007];
  \end{tikzpicture}%
  \hspace{-2mm}%
}
\foreach \x in {A, ..., Z}{\expandafter\xdef\csname orcid\x\endcsname{%
  \noexpand\href{https://orcid.org/\csname orcidauthor\x\endcsname}
  {\noexpand\orcidicon}}%
}

\newcommand{\@preprintleftlogo}{logo-zgca.pdf}
\newcommand{\@preprintrightlogo}{logo-hmilab.pdf}
\newcommand{\preprintlogos}[2]{%
  \gdef\@preprintleftlogo{#1}%
  \gdef\@preprintrightlogo{#2}%
}
\newcommand{\@preprintdate}{May 2026}
\newcommand{\preprintdate}[1]{\gdef\@preprintdate{#1}}
\newcommand{\@preprintfooter}{}
\newcommand{\preprintfooter}[1]{\gdef\@preprintfooter{#1}}
\newcommand{\@runningtitle}{}
\newcommand{\runningtitle}[1]{\gdef\@runningtitle{#1}}
\newcommand{\@printtitle}{\ifthenelse{\equal{\@runningtitle}{}}{\@title}{\@runningtitle}}
\newcommand{\@preprintteaserfile}{}
\newcommand{\@preprintteasercaption}{}
\newcommand{\@preprintteaserlabel}{fig:teaser}
\newcommand{\preprintteaser}[3][fig:teaser]{%
  \gdef\@preprintteaserlabel{#1}%
  \gdef\@preprintteaserfile{#2}%
  \gdef\@preprintteasercaption{#3}%
}
\newcommand{\@preprintprojecturl}{}
\newcommand{\@preprintcodeurl}{}
\newcommand{\@preprintmodelurl}{}
\newcommand{\preprintlinkset}[3]{%
  \gdef\@preprintprojecturl{#1}%
  \gdef\@preprintcodeurl{#2}%
  \gdef\@preprintmodelurl{#3}%
}
\newcommand{\@preprintlinkiconheight}{0.20in}
\newcommand{\@preprintlinkiconwidth}{0.34in}
\newcommand{\@preprintlinks}{%
  \ifthenelse{\equal{\@preprintprojecturl}{}}{}{%
    \vspace*{2pt}%
    \begin{center}%
      \normalsize
      \begin{tabular}{@{}c@{\hspace{0.55em}}l@{}}%
          \makebox[\@preprintlinkiconwidth][c]{%
    \raisebox{\dimexpr.5ex-.5\height\relax}{\includegraphics[height=\@preprintlinkiconheight]{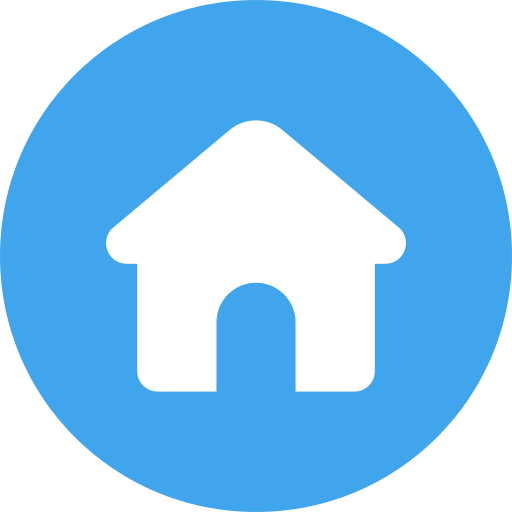}}%
  }%
  &
  \href{\@preprintprojecturl}{\textbf{\texttt{\strut \@preprintprojecturl}}}%
\\[2pt]%
          \makebox[\@preprintlinkiconwidth][c]{%
    \raisebox{\dimexpr.5ex-.5\height\relax}{\includegraphics[height=\@preprintlinkiconheight]{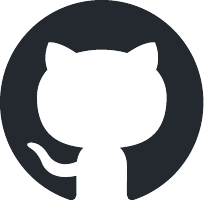}}%
  }%
  &
  \href{\@preprintcodeurl}{\textbf{\texttt{\strut \@preprintcodeurl}}}%
      \end{tabular}%
    \end{center}%
    \vspace*{2pt}%
  }%
}
\newcommand{\@preprintlogoheight}{1.55cm}
\newcommand{\@preprintrulewidth}{7.303in}
\newcommand{\@preprintruleleft}{1.52}
\newcommand{\@preprintdatelength}{5cm}
\newcommand{\@preprintdateleft}{15.07}
\newcommand{\@preprinthfoffsetleft}{0.214in}
\newcommand{\@preprinthfoffsetright}{0.214in}
\newcommand{\@preprintheader}{%
  \begin{textblock}{5}(\@preprintruleleft,0.36)%
    \includegraphics[height=\@preprintlogoheight]{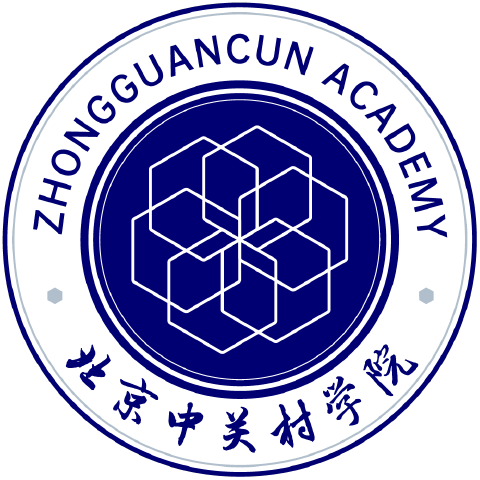}%
  \end{textblock}%
  \begin{textblock}{5}(3.28,0.36)%
    \includegraphics[height=\@preprintlogoheight]{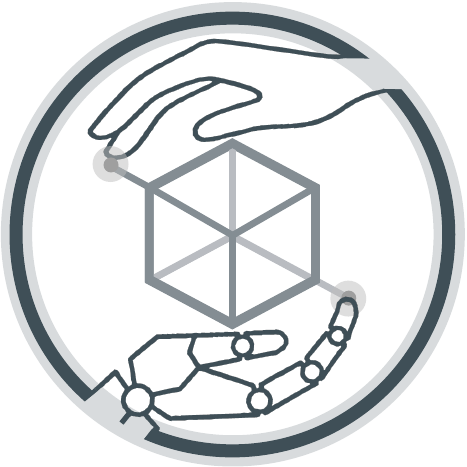}%
  \end{textblock}%
  \begin{textblock}{5}(\@preprintdateleft,1.56)%
    \noindent\makebox[\@preprintdatelength][r]{\footnotesize\@preprintdate}%
  \end{textblock}%
  \begin{textblock}{19}(\@preprintruleleft,2.15)%
    \noindent\rule{\@preprintrulewidth}{0.4pt}%
  \end{textblock}%
  \vspace*{-0.45in}%
}

\fancyhfoffset[L]{\@preprinthfoffsetleft}
\fancyhfoffset[R]{\@preprinthfoffsetright}
\fancypagestyle{preprintfirst}{%
  \fancyhf{}%
  \setlength{\headwidth}{\dimexpr\textwidth+\@preprinthfoffsetleft+\@preprinthfoffsetright\relax}%
  \fancyhfoffset[L]{\@preprinthfoffsetleft}%
  \fancyhfoffset[R]{\@preprinthfoffsetright}%
  \fancyfoot[L]{\footnotesize\itshape \@preprintfooter}%
}

\font\elvbf=ptmb scaled 1100

\newlength{\@ctmp}
\newlength{\@figindent}
\long\def\@makecaption#1#2{%
  \vskip 10pt
  \setbox\@tempboxa\hbox{\normalfont\normalsize\noindent #1.~#2}%
  \setlength{\@ctmp}{\hsize}%
  \addtolength{\@ctmp}{-\@figindent}\addtolength{\@ctmp}{-\@figindent}%
  \ifdim \wd\@tempboxa >\@ctmp
    \begin{list}{}{\leftmargin\@figindent \rightmargin\leftmargin}
      \item[]\normalfont\normalsize #1.~#2\par
    \end{list}
  \else
    \hbox to\hsize{\hfil\box\@tempboxa\hfil}%
  \fi}

\renewcommand{\section}{\@startsection {section}{1}{\z@}
  {14pt plus 2pt minus 2pt}{14pt plus 2pt minus 2pt}{\large\bf}}
\renewcommand{\subsection}{\@startsection {subsection}{2}{\z@}
  {13pt plus 2pt minus 2pt}{13pt plus 2pt minus 2pt}{\elvbf}}

\renewcommand{\maketitle}{%
  \par
  \begingroup
    \renewcommand{\thefootnote}{\fnsymbol{footnote}}%
    \renewcommand{\@makefnmark}{\hbox to \z@{$^{\@thefnmark}$\hss}}%
    \long\def\@makefntext##1{%
      \parindent 1em\noindent
      \hbox to 1.8em{\hss $\m@th ^{\@thefnmark}$}##1}%
    \thispagestyle{preprintfirst}%
    \if@twocolumn
      \twocolumn[\@maketitle]%
    \else
      \@maketitle
    \fi
    \pagestyle{fancy}%
    \@thanks
  \endgroup
  \setcounter{footnote}{0}%
  \let\thanks\relax
}

\renewcommand{\@maketitle}{%
  \null
  \@preprintheader
  \begin{center}
    {\Large\bf \@title \par}
    \vspace*{8pt}
    {\large \lineskip .5em \@author \par}
    \vspace*{4pt}
  \end{center}
  \@preprintlinks
  \@preprintteaser
}

\newcommand{\@preprintteaser}{%
  \ifthenelse{\equal{\@preprintteaserfile}{}}{}{%
    \vspace*{4pt}%
    \begin{center}%
      \includegraphics[width=0.9\textwidth]{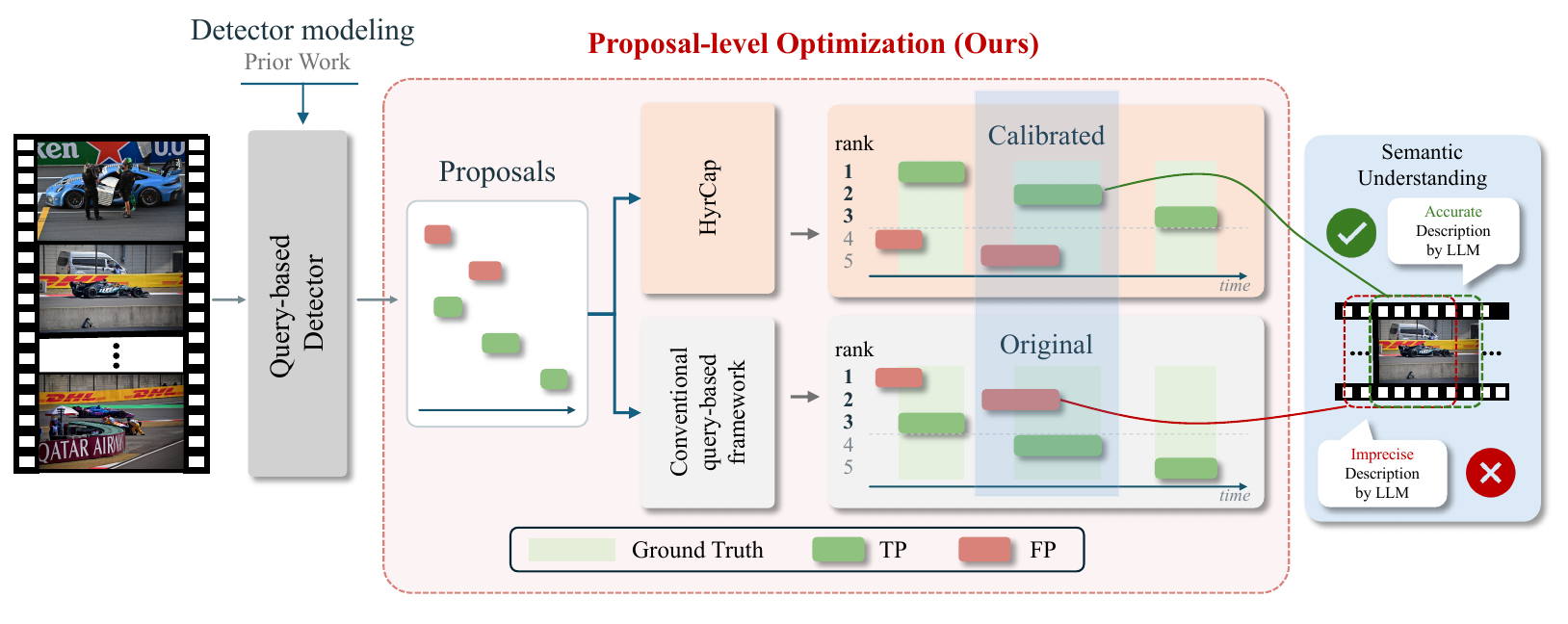}%
    \end{center}%
    \vspace*{-8pt}%
    \begingroup
      \refstepcounter{figure}%
      \@makecaption{\figurename\ \thefigure}{\small \@preprintteasercaption}%
      \label{\@preprintteaserlabel}%
    \endgroup
    \vspace*{6pt}%
  }%
}

\renewenvironment{abstract}
  {%
    \par\begin{center}{\large\bf Abstract}\end{center}%
    \vspace*{6pt}%
    \noindent
  }
  {%
    \par\vspace*{12pt}%
  }

\makeatother

\usepackage{marvosym}
\usepackage{tabularx}
\usepackage{array}
\usepackage{xurl}
\DeclareUnicodeCharacter{2019}{'}
\DeclareUnicodeCharacter{201C}{``}
\DeclareUnicodeCharacter{201D}{''}
\providecommand{\gc}[1]{#1}
\providecommand{\answerYes}[1][]{\textcolor{blue}{[Yes]#1}}
\providecommand{\answerNo}[1][]{\textcolor{orange}{[No]#1}}
\providecommand{\answerNA}[1][]{\textcolor{gray}{[N/A]#1}}
\definecolor{cvprblue}{rgb}{0.21,0.49,0.74}
\hypersetup{
    colorlinks=true, 
    linkcolor=cvprblue, 
    citecolor=cvprblue, 
    urlcolor=cvprblue,  
}

\preprintlogos{logo-zgca.pdf}{logo-hmilab.pdf}

\preprintdate{November 2026}

\runningtitle{HyrCap: Hybrid Rank-Calibration of Action Proposals for  Temporal Event Understanding}

\preprintfooter{\textsuperscript{$\dagger$} Project leader. \textsuperscript{\Letter} Corresponding author.}

\preprintlinkset
  {https://zgca-hmi-lab.github.io/HyrCap}
  {https://github.com/ZGCA-HMI-Lab/HyrCap}
  {}

\preprintteaser{figure-teaser.pdf}{Comparison of HyrCap and conventional query-based TAD: HyrCap produces higher-quality temporal proposals that better support subsequent semantic understanding.}

\title{HyrCap: Hybrid Rank-Calibration of Action Proposals for \\ Temporal Event Understanding}

\author{
Rui Chen\textsuperscript{1} \hspace{0.15in}
Xingyu Chen\textsuperscript{2,\Letter,$\dagger$} \hspace{0.15in}
Pengxin Xu\textsuperscript{4,2} \hspace{0.15in}
Kai Chen\textsuperscript{3,2} \hspace{0.15in}
Junzhi Yu\textsuperscript{1,\Letter} 
}

\affil{
\textsuperscript{1} Peking University \hspace{0.15in} 
\textsuperscript{2} Zhongguancun Academy \\ \textsuperscript{3} Zhongguancun Institute of Artificial Intelligence \hspace{0.15in} 
\textsuperscript{4} Harbin Institute of Technology, Shenzhen
}

\begin{document}

\maketitle

\begin{abstract}
Temporal action detection in long untrimmed videos still suffers from inaccurate temporal localization and low-quality proposals, especially under sparse temporal annotations and ambiguous event boundaries. In the context of open-ended video understanding and LLM-driven event analysis, reliable temporal action proposals have emerged as a critical interface bridging temporal localization, event annotation, and semantic understanding. However, conventional temporal action detection (TAD) pipelines lack explicit modeling of proposal localization quality, cross-representation support, and inter-proposal relations, making it difficult to consistently provide high-quality proposals. We propose Hybrid Rank-Calibration of Action Proposals (HyrCap), a proposal-centric query-based TAD framework for temporal event understanding. In HyrCap, Hybrid denotes the use of complementary video features that provide different temporal localization cues, while Rank-Calibration refers to calibrating the ranking confidence of dense action proposals so that it better reflects proposal localization reliability and candidate competition. HyrCap treats dense query-based predictions as an over-complete temporal event hypothesis space and improves the ranking reliability, localization quality, and candidate discrimination of action proposals through modules such as Cross-Representation Consensus Calibration, Localization Quality Estimation, and Proposal Redundancy Suppression. Furthermore, inspired by the strong capability of LLMs in fine-grained semantic recognition, event description, and reasoning, we introduce HyrCap for Semantic Action Understanding: calibrated proposals can be used for standard TAD detection and as inputs for LLM-based semantic understanding. We further define a staged evaluation protocol to separately measure event localization ability, segment-level semantic understanding, and their combined event understanding performance. Experiments on public benchmarks show that HyrCap achieves state-of-the-art performance across multiple metrics and provides a more fine-grained paradigm and evaluation protocol for query-based event segmentation and annotation. The project page is available at \url{https://zgca-hmi-lab.github.io/HyrCap/}.
\end{abstract}

\section{Introduction}
\label{sec:introduction}

Existing video data typically consist of long untrimmed videos with sparse temporal annotations and imprecise event boundaries, making accurate temporal localization difficult and often leading to low-quality proposals~\cite{idrees2017thumos,caba2015activitynet,zhao2019hacs}.
Localizing reliable event segments from such videos is a fundamental step for video classification, event description, retrieval, and high-level semantic understanding.
Temporal action detection (TAD) aims to localize action instances in untrimmed videos and recognize their semantic categories~\cite{idrees2017thumos,caba2015activitynet}.
Conventional methods usually formulate TAD as a closed-set detection problem, where a model predicts temporal boundaries, assigns action labels, and ranks candidate detections by confidence scores~\cite{lin2019bmn,zhang2022actionformer,shi2023tridet}.
However, in the context of open-ended video understanding and LLM-driven event analysis, reliable temporal action proposals have become a key interface between temporal localization, event annotation, and semantic understanding~\cite{sun2019videobert,maaz2024video,yang2025qwen3}.
Therefore, TAD is expected not only to produce final detections, but also to provide high-quality proposals.

Recent query-based TAD methods generate dense temporal candidates with learnable queries and have shown strong potential for localizing events in long videos~\cite{carion2020end,liu2022end,kim2024te,kim2025digit}.
However, detection performance depends not only on whether the model generates proposals that cover ground-truth events, but also on how these proposals are ranked, suppressed, and matched during evaluation.
Existing pipelines lack explicit modeling of proposal localization quality, cross-representation support, and inter-proposal relations, making dense query predictions with localization potential difficult to preserve, rank, and exploit.
This suggests that improving TAD should not only rely on generating more candidates, but should also understand and optimize how proposals behave throughout the final detection pipeline.

Motivated by this observation, we first revisit the role of proposals in query-based TAD and analyze different video features for this task.
We study how different types of false positives (FPs), proposal ranking, evaluation matching, and video feature representations affect proposal quality and final detection performance~\cite{carreira2017quo,wang2024internvideo2,assran2025v}.
This analysis clarifies the relationship among proposal behavior, feature representation, and final mAP, and indicates that the localization potential of query-based detectors should be released through more reliable proposal-level ranking calibration.

Based on the above analysis, we propose Hybrid Rank-Calibration of Action Proposals (HyrCap), as shown in
Figure~\ref{fig:teaser}, a proposal-centric query-based TAD framework for temporal event understanding. 
In HyrCap, Hybrid denotes the use of complementary video feature representations that provide different temporal localization cues, while Rank-Calibration refers to calibrating the ranking confidence of dense action proposals so that it better reflects proposal localization reliability and candidate competition. 
HyrCap does not directly treat dense query predictions as final detections. 
Instead, it regards them as an over-complete temporal event hypothesis space and improves the ranking reliability, localization quality, and candidate discrimination of action proposals through modules such as Cross-Representation Consensus Calibration, Localization Quality Estimation, and Proposal Duplication Suppression.

Furthermore, inspired by the growing capability of LLMs in fine-grained event description, semantic recognition, and spatio-temporal reasoning~\cite{sun2019videobert,maaz2024video,yang2025qwen3}, we introduce HyrCap for Semantic Action Understanding.
This paradigm argues that query-based TAD should not be assessed only by overall mAP under conventional one-stage loc-cls evaluation, but further decomposed into proposal localization ability and segment-level semantic understanding ability.
Specifically, in the first stage, action categories are collapsed into foreground event and background, and the category-agnostic localization quality, ranking reliability, and event coverage of proposals are evaluated independently.
In the second stage, an LLM or semantic classifier is applied to localized segments to evaluate classification, description, or reasoning performance.
This two-stage protocol decouples whether an event is accurately localized from whether its semantics are correctly understood, making reliable generation, ranking, and selection of temporal action proposals the central capability of query-based detectors.

Our contributions are summarized as follows:

\begin{itemize}
    \item 
    We provide a systematic analysis of proposals in query-based TAD, studying how FPs, proposal ranking, evaluation matching, and different video features affect proposal quality and final detection performance, thereby clarifying the relationship among proposal behavior, feature representation, and final mAP.

    \item 
    We propose HyrCap, a proposal-centric rank-calibration framework that improves proposal localization reliability, ranking quality, and candidate discrimination through modules such as Cross-Representation Consensus Calibration, Localization Quality Estimation, and Proposal Duplication Suppression.

    \item 
    We introduce a two-stage paradigm for semantic action understanding, where proposal localization ability and segment-level semantic understanding ability are evaluated separately. 
    HyrCap achieves state-of-the-art performance across multiple metrics on public benchmarks and provides a more fine-grained paradigm and evaluation protocol for query-based event segmentation and annotation.
\end{itemize}

\section{Related Work}

\paragraph{Video representations for action understanding.}
Action recognition is a fundamental task in video analysis, aiming to classify video sequences into specific action categories. 
I3D \cite{carreira2017quo} extends the Inception network to the video domain by introducing 3D convolutions; R(2+1)D \cite{tran2018closer} improves computational efficiency by decomposing 3D convolutions into 2D spatial operations and 1D temporal operations. 
TSN~\cite{wang2016temporal} divides a video into multiple temporal segments and aggregates segment-level predictions through a consensus function, enabling effective long-range temporal modeling for action recognition.
VideoMAEv2 \cite{wang2023videomae} adopts a Transformer-based masked reconstruction pretraining strategy to learn more robust video representations. 
InternVideo2 \cite{wang2024internvideo2} further enhances video representation learning with large-scale training data and a highly scalable model. 
V-JEPA2 \cite{assran2025v} learns high-level video representations through self-supervised joint embedding prediction, improving spatiotemporal modeling capability. 
These models are commonly used as feature extractors for downstream tasks such as TAD.

\paragraph{Temporal action detection.}
Anchor-free and query-based detectors are two mainstream paradigms in current TAD.
Anchor-free methods such as ActionFormer \cite{zhang2022actionformer}, TriDet \cite{shi2023tridet}, and ActionMamba \cite{chen2026video} directly predict action segments, avoiding predefined anchors and improving temporal modeling with Transformer architectures, trident prediction schemes, or Mamba-based structures, but still typically rely on NMS to remove redundant proposals.
Query-based methods, inspired by DETR~\cite{carion2020end}, formulate TAD as a set prediction problem and generate candidate events with learnable queries, thereby reducing reliance on hand-crafted components.
RTD-Net \cite{tan2021relaxed}, ReAct \cite{shi2022react}, TadTR \cite{liu2022end}, and DualDETR \cite{zhu2024dual} have advanced query-based TAD, but some still depend on sliding windows or NMS.
In contrast, TE-TAD \cite{kim2024te} achieves a complete end-to-end detection framework by redesigning coordinate representation; building on it, DiGIT \cite{kim2025digit} further introduces MDGE, a multi-scale adapter, and CAID to improve encoder representation capability and decoder temporal context modeling.

\paragraph{Localization quality, temporal refinement, and video-language understanding.}
Localization-aware ranking and post-processing have been studied in the 2D object detection domain, including Soft-NMS \cite{bodla2017soft}, IoU-Net~\cite{jiang2018acquisition}, and Weighted Boxes Fusion~\cite{solovyev2021weighted}. 
These methods improve detection from the perspectives of score suppression, localization quality estimation, and box-level fusion. 
However, TAD focuses on temporal action segments, where boundaries are more ambiguous and proposal ranking is affected by redundancy, one-to-one matching, and global score ordering across videos. 
Therefore, post-processing strategies from 2D detection cannot be directly regarded as equivalent solutions for TAD. 
Meanwhile, video-language models such as VideoBERT~\cite{sun2019videobert} show the potential of event-level semantic understanding. 
For query-based TAD, reliable temporal localization is the foundation for subsequent semantic understanding, making high-quality temporal proposals essential for connecting detection with video-language understanding.

\section{Analysis of Proposal Behavior}
\label{sec:analysis}

\begin{figure}
    \centering
    \includegraphics[width=0.9\linewidth]{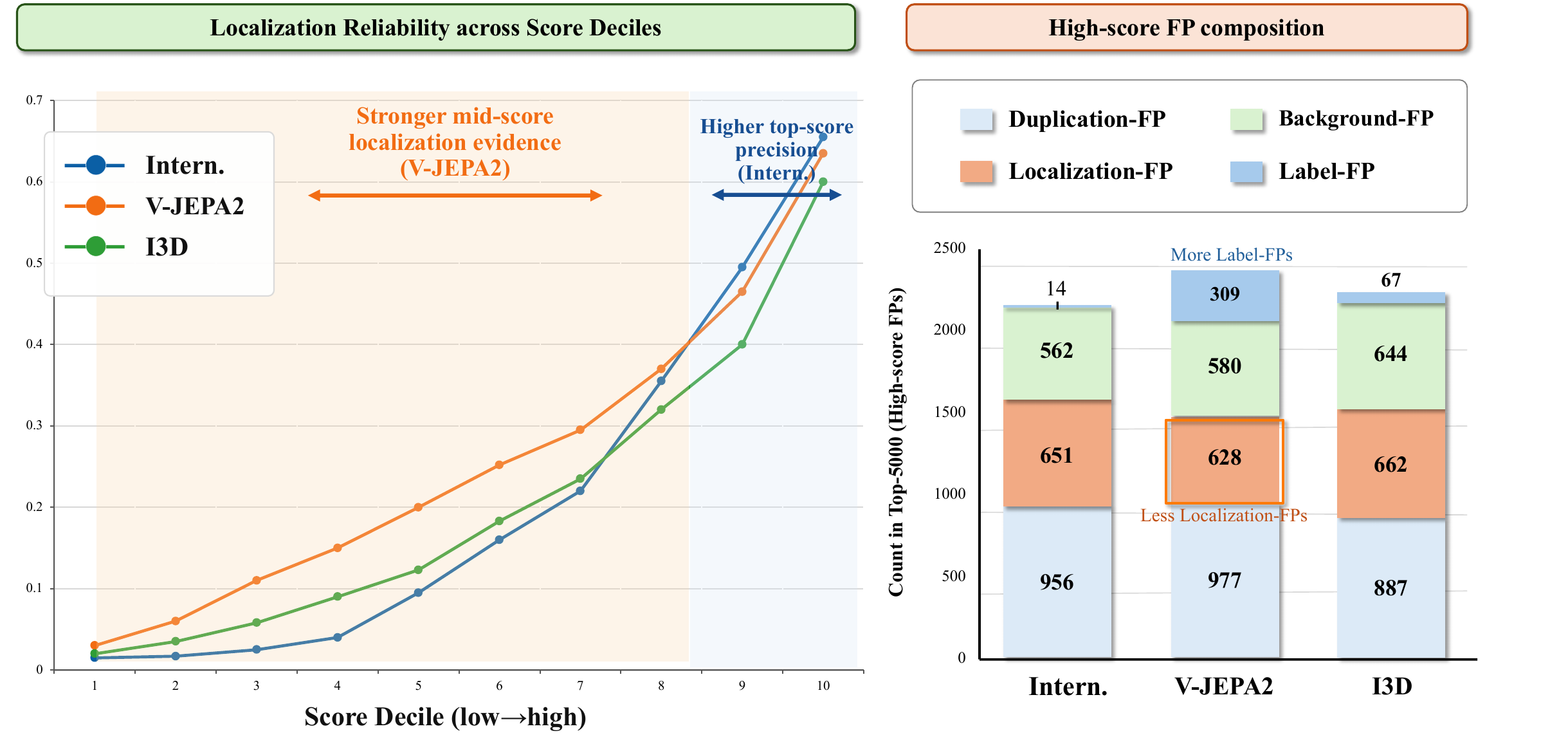}
    \caption{
    Visualization of different features under TAD. Intern. denotes InternVideo2 features.
    }
    \label{fig:ana}

\end{figure}

\paragraph{FP Categorization.}
For dense query predictions, final detection quality is determined not only by the generated candidate segments, but also by temporal boundary accuracy, confidence ranking, proposal redundancy, and one-to-one evaluation matching.
Therefore, how proposals are preserved, ranked, and matched in the final detection pipeline is critical for understanding the performance bottleneck of TAD.
To analyze the sources of false positives (FPs), we instantiate the analysis with DiGIT and compare the proposal behaviors of I3D~\cite{carreira2017quo}, InternVideo2~\cite{wang2024internvideo2}, and V-JEPA2~\cite{assran2025v}.
We categorize FPs into four types: duplication FP denotes redundant proposals caused by one-to-one matching when multiple candidates correspond to the same ground-truth event; localization FP denotes proposals that overlap with a ground-truth event but have inaccurate temporal boundaries; background FP denotes proposals with low overlap to all ground-truth events; and label FP denotes proposals with correct temporal coverage but incorrect class prediction.

\paragraph{Feature-dependent Proposal Behavior.}
As shown in Fig.~\ref{fig:ana}, different feature inputs lead to distinct proposal behaviors.
InternVideo2 shows the strongest proposal quality: it achieves the highest TP rate in the top deciles and produces fewer duplication, background, and label FPs among high-score predictions, indicating cleaner high-confidence candidates and more stable category prediction.
In contrast, although V-JEPA2 is weaker than InternVideo2 as a detection representation, it achieves higher localization reliability in the middle score range and produces fewer localization FPs in the top-5000 predictions, suggesting complementary localization evidence.
I3D exhibits weaker localization reliability and produces more background FPs.
These observations indicate that InternVideo2 is suitable as the primary proposal source, while V-JEPA2 provides complementary localization reliability, motivating the hybrid rank-calibration design of HyrCap.

\section{Method}
\label{sec:method}

The preceding analysis indicates that InternVideo2 and V-JEPA2 exhibit complementary proposal behaviors in TAD. 
Based on this observation, we propose HyrCap, a hybrid rank-calibration framework for query-based temporal action detection. 
As illustrated in Fig.~\ref{fig:method}(a), dense query-based proposals are first generated from the primary InternVideo2 representation and the auxiliary V-JEPA2 representation. 
HyrCap then operates in the proposal space, as shown in Fig.~\ref{fig:method}(b), and recalibrates existing proposal scores by combining learned cross-representation temporal consensus, proposal-level localization reliability, and local proposal competition:
\begin{equation}
    s_i^{cal}
    =
    s_i
    (1+\beta c_i m_i)
    (q_i^{loc})^{\gamma_q}
    (1-d_i)^{\gamma_d},
\end{equation}
where $s_i$ is the detector score, $m_i$ denotes cross-representation temporal support, $c_i$ is the learned consensus gate, $q_i^{loc}$ estimates localization reliability, and $d_i$ is the relation-suppression score.
After calibration, HyrCap refines proposal boundaries by temporal voting and applies multi-class NMS to obtain final detections.

\subsection{Cross-Representation Consensus Calibration}
\label{sec:feature_based_reranking}

\begin{figure}
    \centering
    \includegraphics[width=0.99\linewidth]{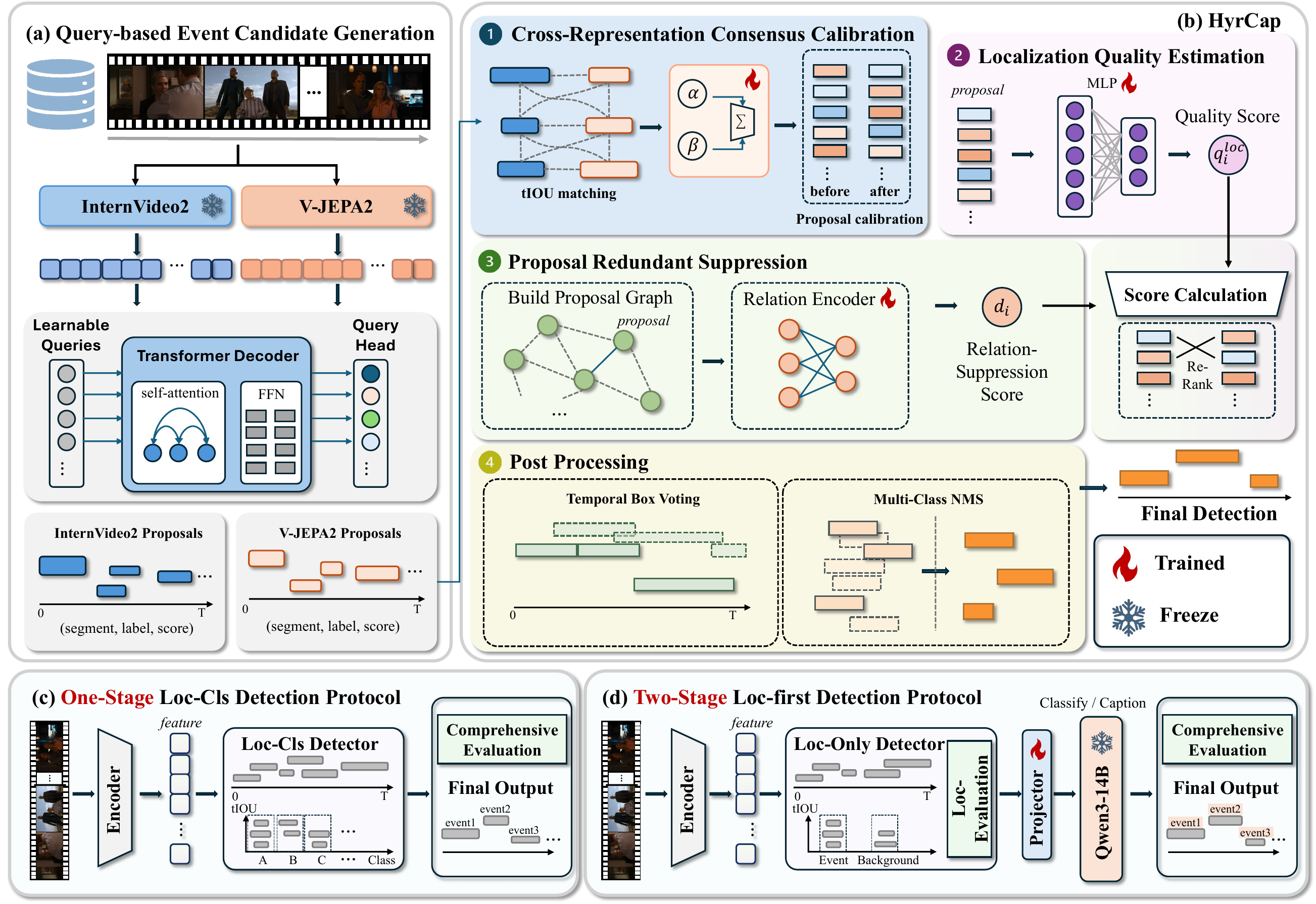}
    \caption{
    Overview of HyrCap.
    (a) Query-based event candidate generation with InternVideo2 and V-JEPA2 representations.
    (b) HyrCap calibrates and refines dense proposals through cross-representation consensus calibration, localization quality estimation, proposal duplication suppression, and post-processing.
    (c) Conventional one-stage loc-cls detection protocol.
    (d) Two-stage loc-first protocol for staged semantic understanding.
    Loc denotes localization.
    }

    \label{fig:method}
\end{figure}

As shown in Fig.~\ref{fig:method}(b)-1, CPCC uses InternVideo2 proposals as the primary candidates and V-JEPA2 proposals as class-agnostic auxiliary temporal evidence.
The primary proposal set is denoted as
$\mathcal{P}^{I}=\{p_i=(b_i,y_i,s_i)\}_{i=1}^{N}$, where $b_i$ is the temporal segment, $y_i$ is the predicted category, and $s_i$ is the detector confidence.
The auxiliary proposal set is denoted as
$\mathcal{P}^{A}=\{\bar{p}_j=(\bar{b}_j)\}_{j=1}^{M}$.

For each primary proposal, we compute its auxiliary temporal support by the maximum tIoU:
\begin{equation}
    m_i=\max_{\bar{p}_j\in\mathcal{P}^{A}}\operatorname{tIoU}(b_i,\bar{b}_j).
\end{equation}
Instead of using $m_i$ as a fixed heuristic, CPCC trains a lightweight consensus head to predict a gate $c_i=f_{\theta}^{con}(r_i)$ from proposal-level consensus cues.
The gate is supervised by the localization quality of the primary proposal:
\begin{equation}
\hat{c}_i =
\begin{cases}
1, & \max_{g\in\mathcal{G}_v}
\operatorname{tIoU}(b_i,b_g) \geq \mu_c,\\
0, & \text{otherwise},
\end{cases}
\quad
\mathcal{L}_{con} =
\operatorname{BCE}(c_i,\hat{c}_i).
\end{equation}
The consensus-calibrated score is defined as
$s_i^{con}=s_i(1+\beta c_i m_i)$.
This module keeps the original segment and category unchanged, and only adjusts proposal ranking through learned cross-representation temporal consensus.

\subsection{Localization Quality Estimation}
\label{sec:localization_based_reranking}

Fig.~\ref{fig:method}(b)-2 shows the localization quality estimation module. 
Detector confidence does not always reflect temporal localization quality. 
HyrCap therefore estimates a proposal-level localization score $q_i^{loc}$ from a compact descriptor $x_i^{loc}$, which includes confidence statistics, temporal geometry, consensus evidence, and local ranking cues:
\begin{equation}
    q_i^{loc}=f_{\theta}^{loc}(x_i^{loc}), \quad q_i^{loc}\in[0,1].
\end{equation}
The supervision is constructed offline from ground-truth annotations.
A proposal is localization-positive when it sufficiently overlaps with any ground-truth event in the same video:
\begin{equation}
    \mathcal{L}_{loc}
    =
    \operatorname{BCE}(q_i^{loc},\hat{q}_i),
    \quad
    \hat{q}_i
    =
    \begin{cases}
    1, & \max_{g\in\mathcal{G}_v}\operatorname{tIoU}(b_i,b_g)\geq\mu_q,\\
    0, & \mathrm{otherwise}.
    \end{cases}
\end{equation}
This target is category-agnostic and measures temporal alignment rather than semantic correctness. 
During inference, $q_i^{loc}$ is used as a multiplicative factor in the calibrated ranking score.

\subsection{Proposal Duplication Suppression}
\label{sec:relation_based_reranking}

As illustrated in Fig.~\ref{fig:method}(b)-3, dense query predictions often contain multiple overlapping proposals around the same event. 
To model local proposal competition, HyrCap builds same-video and same-class neighborhoods according to temporal overlap. 
For each proposal, pairwise cues such as temporal IoU, center distance, length ratio, and score gap are aggregated by a lightweight relation encoder:
\begin{equation}
    d_i=f_{\phi}^{rel}(h_i'), \quad d_i\in[0,1].
\end{equation}
Here, $d_i$ estimates whether proposal $p_i$ should be suppressed under its local proposal context. 
The suppression target is constructed from offline proposal matching diagnosis, where duplicate proposals and locally inferior candidates in overlapping groups are treated as suppression-positive samples:
\begin{equation}
    \mathcal{L}_{cal}
    =
    \mathcal{L}_{loc}
    +
    \lambda_{rel}\mathcal{L}_{rel},
    \quad
    \mathcal{L}_{rel}
    =
    \operatorname{BCE}(d_i,\hat{d}_i).
\end{equation}
The relation factor $(1-d_i)^{\gamma_d}$ reduces the ranking priority of redundant or inferior candidates while preserving proposals with stronger local evidence.

After reranking, HyrCap further refines proposal boundaries by temporal voting, as shown in Fig.~\ref{fig:method}(b)-4. 
For each proposal $p_i$, we collect its same-video and same-class neighbors $\mathcal{B}_i$ whose tIoU with $p_i$ exceeds a threshold $c$. 
The refined boundaries are computed as
\begin{equation}
    \hat{t}_i^s
    =
    \frac{\sum_{p_j\in\mathcal{B}_i} w_j t_j^s}{\sum_{p_j\in\mathcal{B}_i}w_j+\epsilon},
    \quad
    \hat{t}_i^e
    =
    \frac{\sum_{p_j\in\mathcal{B}_i} w_j t_j^e}{\sum_{p_j\in\mathcal{B}_i}w_j+\epsilon},
    \quad
    w_j=\max(s_j^{cal}-\tau,0).
\end{equation}
The final refined proposal is $\hat{p}_i=(\hat{b}_i,y_i,s_i^{cal})$, where $\hat{b}_i=(\hat{t}_i^s,\hat{t}_i^e)$. 
Multi-class NMS is then applied to produce the final detection results.

\subsection{HyrCap for Staged Semantic Understanding}
\label{sec:staged_semantic_evaluation}

Beyond closed-set TAD evaluation, HyrCap supports a staged semantic understanding protocol.
As shown in Fig.~\ref{fig:method}(c), the conventional one-stage loc-cls protocol jointly performs temporal localization and closed-set classification, where proposal ranking, boundary quality, and semantic prediction are coupled in the final class-specific detections.

In contrast, Fig.~\ref{fig:method}(d) shows our two-stage loc-first protocol.
In the first stage, HyrCap focuses on category-agnostic event localization and converts dense query predictions into a compact set of event proposals through rank calibration, boundary refinement, and class-agnostic NMS:
\begin{equation}
    \mathcal{E}(V)
    =
    \operatorname{TopK}
    \left(
    \operatorname{NMS}_{ca}
    \left(
    \{(\hat{b}_i,s_i^{cal})\}_{i=1}^{N}
    \right)
    \right),
\end{equation}
where $\mathcal{E}(V)$ denotes the selected event units and $\operatorname{NMS}_{ca}$ denotes class-agnostic NMS.
The selection only depends on the calibrated localization score $s_i^{cal}$, without using ground-truth annotations or semantic model confidence.

In the second stage, each selected event segment is passed to a semantic model through a trained projector for segment-level classification, captioning, or reasoning:
\begin{equation}
    (\tilde{y}_i,\tilde{r}_i)
    =
    \mathcal{M}_{sem}
    \left(
    g_{\omega}\left(\psi(V[\hat{b}_i])\right),
    \mathcal{C},
    \pi
    \right),
\end{equation}
where $\psi(V[\hat{b}_i])$ denotes the segment-level visual representation, $g_{\omega}(\cdot)$ is the trained projector, $\mathcal{C}$ is the candidate action vocabulary, and $\pi$ is the task prompt.
The final staged output is
\begin{equation}
    \mathcal{D}_{stage}(V)
    =
    \{(\hat{b}_i,\tilde{y}_i,s_i^{cal},\tilde{r}_i)\}_{i=1}^{K}.
\end{equation}
The ranking score remains $s_i^{cal}$ rather than the semantic model confidence, keeping temporal proposal ranking and semantic interpretation decoupled.
For joint evaluation, a prediction is considered correct only when both its temporal segment and semantic prediction are correct.

\begin{table*}[t]
\centering
\caption{Comparison with state-of-the-art methods on THUMOS14 \cite{idrees2017thumos} and ActivityNet v1.3 \cite{caba2015activitynet}. For TE
-TAD + Ours, we additionally apply our methods on TE-TAD.}
\label{tab:sota_thumos_activitynet}
\small
\renewcommand{\arraystretch}{1.28}
\setlength{\tabcolsep}{4.2pt}
\setlength{\extrarowheight}{0.8pt}
\resizebox{\textwidth}{!}{
\begin{tabular}{c c l l c c c c c c c c c c}
\toprule
\multirow{2}{*}{\makecell[c]{Training\\Type}} 
& \multirow{2}{*}{\makecell[c]{Head\\Type}} 
& \multirow{2}{*}{Method} 
& \multirow{2}{*}{Feature}
& \multicolumn{6}{c}{THUMOS14}
& \multicolumn{4}{c}{ActivityNet v1.3} \\
\cmidrule(lr){5-10} \cmidrule(lr){11-14}
& & & 
& 0.3 & 0.4 & 0.5 & 0.6 & 0.7 & Avg.
& 0.5 & 0.75 & 0.95 & Avg. \\
\midrule

\multirow{4}{*}{Full}
& \multirow{4}{*}{-}
& AFSD & I3D \cite{carreira2017quo}
& 67.3 & 62.4 & 55.5 & 43.7 & 31.1 & 52.0
& 52.4 & 35.3 & 6.5 & 34.4 \\
& & TALLFormer & Swin-B \cite{liu2021swin}
& 76.0 & - & 63.2 & - & 34.5 & 59.2
& 54.1 & 36.2 & 7.9 & 35.6 \\
& & ViT-TAD & ViT-B \cite{dosovitskiy2020image}
& 85.1 & 80.9 & 74.2 & 61.8 & 45.4 & 69.5
& 55.8 & 38.5 & 8.8 & 37.4 \\
& & AdaTAD & VideoMAEv2-g \cite{wang2023videomae}
& \best{89.7} & \best{86.7} & \best{80.9} & \best{71.0} & \best{56.1} & \best{76.9}
& \best{61.7} & \best{43.4} & \best{10.9} & \best{41.9} \\
\midrule

\multirow{17}{*}{\makecell[c]{Head\\only}}
& \multirow{5}{*}{\makecell[c]{Anchor\\-free}}
& TriDet & I3D \cite{carreira2017quo} / R(2+1)D \cite{tran2018closer}
& 83.6 & 80.1 & 72.9 & 62.4 & 47.4 & 69.3
& 54.7 & 38.0 & 8.4 & 36.8 \\
& & DyFADet & I3D \cite{carreira2017quo} / R(2+1)D \cite{tran2018closer}
& 84.0 & 80.1 & 72.7 & 61.1 & 47.9 & 69.2
& 58.1 & 39.6 & 8.4 & 38.5 \\
& & ActionFormer & I3D \cite{carreira2017quo} / R(2+1)D \cite{tran2018closer}
& 82.1 & 77.8 & 71.0 & 59.4 & 43.9 & 66.8
& 54.7 & 37.8 & 8.4 & 36.6 \\
& & ActionFormer & InternVideo2 \cite{wang2024internvideo2}
& 82.3 & 81.9 & 75.1 & 65.8 & 50.3 & 71.1
& 61.5 & \best{44.6} & \best{12.7} & 41.2 \\
& & ActionMamba & InternVideo2 \cite{wang2024internvideo2}
& \best{86.9} & \best{83.1} & \best{76.9} & \best{65.9} & \best{50.8} & \best{72.7}
& \best{62.4} & 43.5 & 10.2 & \best{42.0} \\
\cmidrule(lr){2-14}

& \multirow{12}{*}{\makecell[c]{Query\\-based}}
& RTD-Net & I3D \cite{carreira2017quo} / TSN \cite{wang2016temporal}
& 68.3 & 62.3 & 51.9 & 38.8 & 23.7 & 49.0
& 47.2 & 30.7 & 8.6 & 30.8 \\
& & ReAct & I3D \cite{carreira2017quo} / TSN \cite{wang2016temporal}
& 69.2 & 65.0 & 57.1 & 47.8 & 35.6 & 55.0
& 49.6 & 33.0 & 8.6 & 32.6 \\
& & Self-DETR & I3D \cite{carreira2017quo}
& 74.6 & 69.5 & 60.0 & 47.6 & 31.8 & 56.7
& 52.3 & 33.7 & 8.4 & 33.8 \\
& & TadTR & I3D \cite{carreira2017quo} / R(2+1)D \cite{tran2018closer}
& 74.8 & 69.1 & 60.1 & 46.6 & 32.8 & 56.7
& 53.6 & 37.5 & 10.6 & 36.8 \\
& & DualDETR & I3D \cite{carreira2017quo}
& 82.9 & 78.0 & 70.4 & 58.5 & 44.4 & 66.8
& 52.6 & 35.0 & 7.8 & 34.3 \\
& & TE-TAD & I3D \cite{carreira2017quo} / R(2+1)D \cite{tran2018closer}
& 83.3 & 78.4 & 71.3 & 60.7 & 45.6 & 67.9
& 54.2 & 38.1 & 10.6 & 37.1 \\

& & \gc{DiGIT} & \gc{I3D \cite{carreira2017quo} / R(2+1)D} \cite{tran2018closer}
& \gc{83.6} & \gc{79.6} & \gc{71.9} & \gc{61.5} & \gc{48.6} & \gc{69.0}
& \gc{54.4} & \gc{38.2} & \gc{10.7} & \gc{37.3} \\
\cdashline{3-14}[0.8pt/2pt]

& & \gc{TadTR} & \gc{InternVideo2 \cite{wang2024internvideo2}}
& \gc{84.8} & \gc{79.3} & \gc{70.4} & \gc{58.2} & \gc{43.8} & \gc{67.3}
& \gc{57.1} & \gc{38.8} & \gc{11.0} & \gc{38.2} \\
& & TE-TAD  & \gc{InternVideo2 \cite{wang2024internvideo2}}
& \gc{84.3} & \gc{81.1} & \gc{73.7} & \gc{62.6} & \gc{49.5} & \gc{70.3}
& \gc{61.3} & \gc{41.8} & \gc{10.9} & \gc{41.1} \\

& & \gc{DiGIT} & \gc{InternVideo2 \cite{wang2024internvideo2}}
& \gc{87.6} & \gc{84.2} & \gc{77.6} & \gc{67.3} & \gc{52.5} & \gc{73.8}
& \gc{62.0} & \gc{43.1} & \gc{11.3} & \gc{42.0} \\
\cdashline{3-14}[0.8pt/2pt]
& & \gc{TE-TAD + Ours} & \gc{InternVideo2 \cite{wang2024internvideo2} + V-JEPA2 \cite{assran2025v}}
& 85.9 & 81.4 & 74.1 & 66.9 & 48.6 & 71.4
& 62.1 & 42.5 & 11.0 & 41.9 \\

& & \gc{HyrCap} & \gc{InternVideo2 \cite{wang2024internvideo2} + V-JEPA2 \cite{assran2025v}}
& \gc{\best{88.1}} & \gc{\best{85.2}} & \gc{\best{79.1}} & \gc{\best{70.8}} & \gc{\best{54.6}} & \gc{\best{75.6}}
& \gc{\best{64.4}} & \gc{\best{44.0}} & \gc{\best{11.4}} & \gc{\best{43.3}} \\

\bottomrule
\end{tabular}
}

\end{table*}

\section{Experiment}

\subsection{Setup}

\paragraph{Datasets.}
We conduct experiments on three public TAD datasets: THUMOS14~\cite{idrees2017thumos}, ActivityNet v1.3~\cite{caba2015activitynet}, and HACS-Segment~\cite{zhao2019hacs}. THUMOS14 contains 20 action classes with 200/213 validation/test videos and 3,007/3,358 action instances. ActivityNet v1.3 and HACS-Segment both cover 200 classes, with 10,024/4,926 and 37,613/5,981 training/validation videos, respectively.

\paragraph{Evaluation Metrics.}
We follow the standard evaluation protocols of each dataset. 
For THUMOS14, we report mAP at tIoU thresholds of 0.3, 0.4, 0.5, 0.6, and 0.7, together with the average mAP. 
For ActivityNet v1.3 and HACS-Segment, we report mAP@0.5, mAP@0.75, mAP@0.95, and the average mAP over 0.5:0.95. 
In addition to standard class-specific detection evaluation, we report category-agnostic localization results. 
For staged semantic evaluation, we further measure event localization, segment-level semantic understanding, and their combined performance.

\paragraph{Implementation Details.}
For better visualization and analysis, we use DiGIT as the baseline and deploy our method within its framework. More implementation details are provided in the appendix.

\subsection{Comparison with State-of-the-Art Methods}

We first compare HyrCap with existing state-of-the-art temporal action detection methods. 
The compared methods include anchor-based, anchor-free, and query-based detectors. 
The main results are reported on multiple public benchmarks to evaluate the effectiveness of HyrCap under different dataset scales, action category distributions, and video characteristics.

Table \ref{tab:sota_thumos_activitynet} reports the comparison on THUMOS14 and ActivityNet v1.3. 
On THUMOS14, HyrCap improves over strong query-based baselines in terms of average mAP and remains effective under stricter tIoU thresholds. 
This indicates that hybrid rank-calibration not only improves proposal ranking, but also preserves proposals with more accurate temporal boundaries. 
On ActivityNet v1.3, HyrCap also achieves consistent improvement, showing that the proposal-level calibration strategy is not limited to short action instances or a single benchmark.

Table \ref{tab:hacs} reports the results on HACS-Segment.
The performance on this large-scale benchmark shows that HyrCap generalizes to diverse action categories and video distributions.
These results validate HyrCap as a general proposal-centric framework for improving query-based TAD.

\begin{table}[t]
\centering
\caption{Performance comparison with state-of-the-art methods on HACS-Segment.}
\label{tab:hacs}
\normalsize
\renewcommand{\arraystretch}{0.99}
\setlength{\tabcolsep}{3.0pt}
\scalebox{1.0}{
\begin{tabular}{c l l c c c c}
\toprule
\multirow{2}{*}{\mbox{Head Type}}
& \multirow{2}{*}{\mbox{Method}}
& \multirow{2}{*}{\mbox{Feature}}
& \multicolumn{4}{c}{mAP} \\
\cmidrule(lr){4-7}
& & & 0.5 & 0.75 & 0.95 & Avg. \\
\midrule

\multirow{4}{*}{Anchor-based}
& SSN \cite{zhao2017temporal} & I3D \cite{carreira2017quo}
& 28.8 & 18.8 & 5.3 & 19.0 \\
& G-TAD \cite{xu2020g} & I3D \cite{carreira2017quo}
& 41.1 & 27.6 & 8.3 & 27.5 \\
& BMN \cite{lin2019bmn} & SlowFast \cite{feichtenhofer2019slowfast}
& 52.5 & 36.4 & 10.4 & 35.8 \\
& TCANet \cite{qing2021temporal} & SlowFast \cite{feichtenhofer2019slowfast}
& 54.1 & 37.2 & 11.3 & 36.8 \\

\midrule

\multirow{8}{*}{Anchor-free}
& TALLFormer \cite{cheng2022tallformer} & Swin-B \cite{liu2021swin}
& 55.0 & 36.1 & 11.8 & 36.5 \\
& TriDet \cite{shi2023tridet} & I3D \cite{carreira2017quo}
& 54.5 & 36.8 & 11.5 & 36.8 \\
& TriDet \cite{shi2023tridet} & SlowFast \cite{feichtenhofer2019slowfast}
& 56.7 & 39.3 & 11.7 & 38.6 \\
& TriDet \cite{shi2023tridet} & VideoMAEv2-g \cite{wang2023videomae}
& 62.4 & 44.1 & 13.1 & 43.1 \\
& DyFADet \cite{yang2024dyfadet} & SlowFast \cite{feichtenhofer2019slowfast}
& 57.8 & 39.8 & 11.8 & 39.2 \\
& DyFADet \cite{yang2024dyfadet} & VideoMAEv2-g \cite{wang2023videomae}
& 64.0 & 44.8 & 14.1 & 44.3 \\
& ActionFormer \cite{zhang2022actionformer} & InternVideo2 \cite{wang2024internvideo2} 
& 62.6 & 44.6 & 12.7 & 43.3 \\
& ActionMamba \cite{chen2026video} & InternVideo2 \cite{wang2024internvideo2}
& {\bfseries 64.0} & {\bfseries 45.7} & {\bfseries 13.3} & {\bfseries 44.6} \\

\midrule

\multirow{5}{*}{Query-based}
& TadTR \cite{liu2022end} & I3D \cite{carreira2017quo}
& 47.1 & 32.1 & 10.9 & 32.1 \\
& TadTR \cite{liu2022end} & InternVideo2 \cite{wang2024internvideo2}
& 54.2 & 38.8 & 12.8 & 37.8 \\
& TE-TAD \cite{kim2024te} & InternVideo2 \cite{wang2024internvideo2}
& 60.4 & 45.6 & 16.5 & 44.1 \\
& DiGIT \cite{kim2025digit} & InternVideo2 \cite{wang2024internvideo2}
& 61.1 & 47.5 & 17.8 & 45.5 \\
& HyrCap & InternVideo2 \cite{wang2024internvideo2} + V-JEPA2     \cite{assran2025v}
& {\bfseries 62.7} & {\bfseries 48.8} & {\bfseries 17.9} & {\bfseries 47.1} \\

\bottomrule
\end{tabular}
}
\end{table}

\subsection{Ablation Study}

To validate the effectiveness of each component, we conduct ablation experiments on each module, as shown in Table~\ref{tab:ablation}.
CPCC improves proposal ranking by exploiting cross-representation temporal consistency, LQE explicitly estimates proposal localization reliability and prioritizes candidates with more accurate temporal boundaries, and PRS suppresses redundant candidates to reduce high-score false positives.
When combined, these modules achieve the best performance on different baselines.
The Top-100 FP statistics further show that HyrCap effectively reduces duplicate, background, and localization FPs, leading to more reliable proposal ranking and improved detection performance.

\begin{table*}[t]
\centering
\caption{Component Analysis. CPCC denotes Cross-Representation Consensus Calibration, LQE denotes Localization Quality Estimation, and PRS denotes Proposal Redundancy Suppression. Dup. represents Duplication FPs, Bg. Background FPs, Loc. Localization FPs, and Lab. Label FPs.}
\label{tab:ablation}

\def\tableScale{0.9}              
\renewcommand{\arraystretch}{1.15} 
\setlength{\tabcolsep}{5.0pt}      

\resizebox{\tableScale\textwidth}{!}{%
\begin{tabular}{l c c c c c c c c c c c c c}
\toprule
\multirow{2}{*}{Baseline}
& \multicolumn{2}{c}{Input Feature}
& \multirow{2}{*}{CPCC}
& \multirow{2}{*}{LQE}
& \multirow{2}{*}{PRS}
& \multicolumn{4}{c}{mAP}
& \multicolumn{4}{c}{FPs in Top-100} \\
\cmidrule(lr){2-3}
\cmidrule(lr){7-10}
\cmidrule(lr){11-14}
& Intern.
& V-JEPA2
& & &
& 0.3 & 0.5 & 0.7 & Avg.
& Dup. & Bg. & Loc. & Lab. \\
\midrule

\multirow{9}{*}{TE-TAD \cite{kim2024te}}
& $\checkmark$ &
& & &
& 85.0 & 73.0 & 46.5 & 70.3
& 2 & 2 & 1 & 1 \\

& & $\checkmark$
& & &
& 67.5 & 56.7 & 38.4 & 55.2
& 0 & 1 & 1 & 0 \\

& $\checkmark$ & $\checkmark$
& $\checkmark$ & &
& 85.2 & 73.8 & 48.1 & 71.1
& 0 & 1 & 1 & 0 \\

& $\checkmark$ &
& & $\checkmark$ &
& 85.0 & 73.0 & 47.2 & 70.9
& 2 & 2 & 0 & 0 \\

& $\checkmark$ &
& & & $\checkmark$
& 85.2 & 73.1 & 46.6 & 70.7
& 0 & 4 & 1 & 1 \\

& $\checkmark$ & $\checkmark$
& $\checkmark$ & $\checkmark$ &
& 85.0 & 73.0 & 47.2 & 70.8
& 0 & 2 & 0 & 1 \\

& $\checkmark$ & $\checkmark$
& $\checkmark$ & & $\checkmark$
& 85.6 & 73.9 & 48.3 & 70.4
& 0 & 3 & 0 & 2 \\

& $\checkmark$ &
& & $\checkmark$ & $\checkmark$
& 85.0 & 73.4 & 48.0 & 71.0
& 0 & 4 & 1 & 1 \\

& $\checkmark$ & $\checkmark$
& $\checkmark$ & $\checkmark$ & $\checkmark$
& \textbf{85.9}
& \textbf{74.1}
& \textbf{48.6}
& \textbf{71.4}
& \textbf{0}
& \textbf{1}
& \textbf{0}
& \textbf{0} \\

\midrule

\multirow{9}{*}{DiGIT \cite{kim2025digit}}
& $\checkmark$ &
& & &
& 86.6 & 77.3 & 52.8 & 73.8
& 2 & 7 & 6 & 0 \\

& & $\checkmark$
& & &
& 73.9 & 64.3 & 40.9 & 60.8
& 1 & 1 & 0 & 2 \\

& $\checkmark$ & $\checkmark$
& $\checkmark$ & &
& 86.9 & 77.7 & 54.4 & 75.0
& 1 & 4 & 1 & 0 \\

& $\checkmark$ &
& & $\checkmark$ &
& 86.8 & 77.4 & 53.7 & 74.5
& 2 & 6 & 2 & 0 \\

& $\checkmark$ &
& & & $\checkmark$
& 87.7 & 77.8 & 52.8 & 74.7
& 1 & 5 & 5 & 0 \\

& $\checkmark$ & $\checkmark$
& $\checkmark$ & $\checkmark$ &
& 87.3 & 77.8 & 54.5 & 75.1
& 2 & 5 & 1 & 0 \\

& $\checkmark$ & $\checkmark$
& $\checkmark$ & & $\checkmark$
& 87.7 & 78.5 & 54.2 & 75.4
& 0 & 2 & 4 & 0 \\

& $\checkmark$ &
& & $\checkmark$ & $\checkmark$
& 87.6 & 77.7 & 52.2 & 74.8
& 0 & 5 & 6 & 0 \\

& $\checkmark$ & $\checkmark$
& $\checkmark$ & $\checkmark$ & $\checkmark$
& \textbf{88.1}
& \textbf{79.1}
& \textbf{54.6}
& \textbf{75.6}
& \textbf{0}
& \textbf{1}
& \textbf{1}
& \textbf{0} \\

\bottomrule
\end{tabular}%
}

\end{table*}

\paragraph{Controlled Comparison of Fusion and Calibration Strategies.}
Because CPCC introduces proposals from an auxiliary representation, we further conduct controlled comparisons to distinguish the benefit of additional proposals from that of the proposed calibration mechanism. All methods use exactly the same InternVideo2 primary proposals and V-JEPA2 auxiliary proposals, together with the same proposal budget, score normalization, Top-$K$ setting, NMS threshold, and evaluation protocol. Direct Union + NMS directly merges proposals from both branches before NMS; Score Average and Max-Score Fusion combine the scores of matched proposals; Temporal WBF fuses their temporal boundaries using proposal-score-based weights; and Fixed Consensus uniformly sets the consensus gate to $c_i=1$. Therefore, the performance differences reflect only the fusion, calibration, and suppression strategies.

\begin{table}[t]
\centering
\caption{Controlled comparison of fusion and calibration strategies on THUMOS14 using identical dual-branch proposal inputs.}
\label{tab:controlled_comparison}

\def\tableScale{0.55}              
\renewcommand{\arraystretch}{1.15} 
\setlength{\tabcolsep}{5.0pt}      

\resizebox{\tableScale\columnwidth}{!}{%
\begin{tabular}{l c c c c}
\toprule
Method
& AP@0.3
& AP@0.5
& AP@0.7
& Avg. mAP \\
\midrule

Direct Union + NMS
& 80.9
& 72.2
& 48.2
& 68.2 \\

Score Average
& 81.8
& 72.6
& 48.4
& 68.6 \\

Max-Score Fusion
& 80.5
& 71.1
& 45.7
& 66.8 \\

Temporal WBF
& 83.8
& 75.2
& 50.8
& 71.1 \\

Fixed Consensus
& 87.3
& 78.0
& 53.8
& 74.1 \\

HyrCap
& \textbf{88.1}
& \textbf{79.1}
& \textbf{54.6}
& \textbf{75.6} \\

\bottomrule
\end{tabular}%
}

\end{table}

As shown in Table~\ref{tab:controlled_comparison}, directly combining the two proposal sources or applying simple score-level fusion remains substantially inferior to HyrCap, showing that the improvement cannot be attributed merely to the availability of additional proposals. Temporal WBF improves over simple fusion, while Fixed Consensus further confirms that cross-branch temporal agreement provides useful reliability evidence. HyrCap surpasses Fixed Consensus by 1.5 points in average mAP, demonstrating that learned proposal-level calibration can selectively exploit beneficial auxiliary evidence while limiting the influence of unreliable cross-branch agreement.

\subsection{Staged Semantic Evaluation}

Beyond closed-set TAD evaluation, we further evaluate HyrCap under a staged semantic understanding protocol on THUMOS14 with InternVideo2 features.
This protocol first evaluates whether a detector can produce category-agnostic event proposals, and then sends the selected event segments to Qwen3-14B~\cite{yang2025qwen3} for segment-level classification or captioning.

Specifically, for localization evaluation, all action categories are collapsed into foreground event and background, and proposals are ranked by localization scores and matched to ground-truth segments using standard tIoU-based AP computation.
For semantic evaluation, we select the top-ranked proposals after class-agnostic NMS, extract their segment features, map them with the trained projector, and feed them into Qwen3-14B for segment-level semantic prediction.
For joint event evaluation, proposals are ranked by the calibrated localization score, and a prediction is counted as a true positive only when its temporal segment matches an unmatched ground-truth instance under the required tIoU threshold and its semantic prediction is correct. Thus, semantic confidence is not used for proposal ranking, keeping temporal localization and semantic interpretation decoupled.

\begin{figure}
    \centering
    \includegraphics[width=0.85\linewidth]{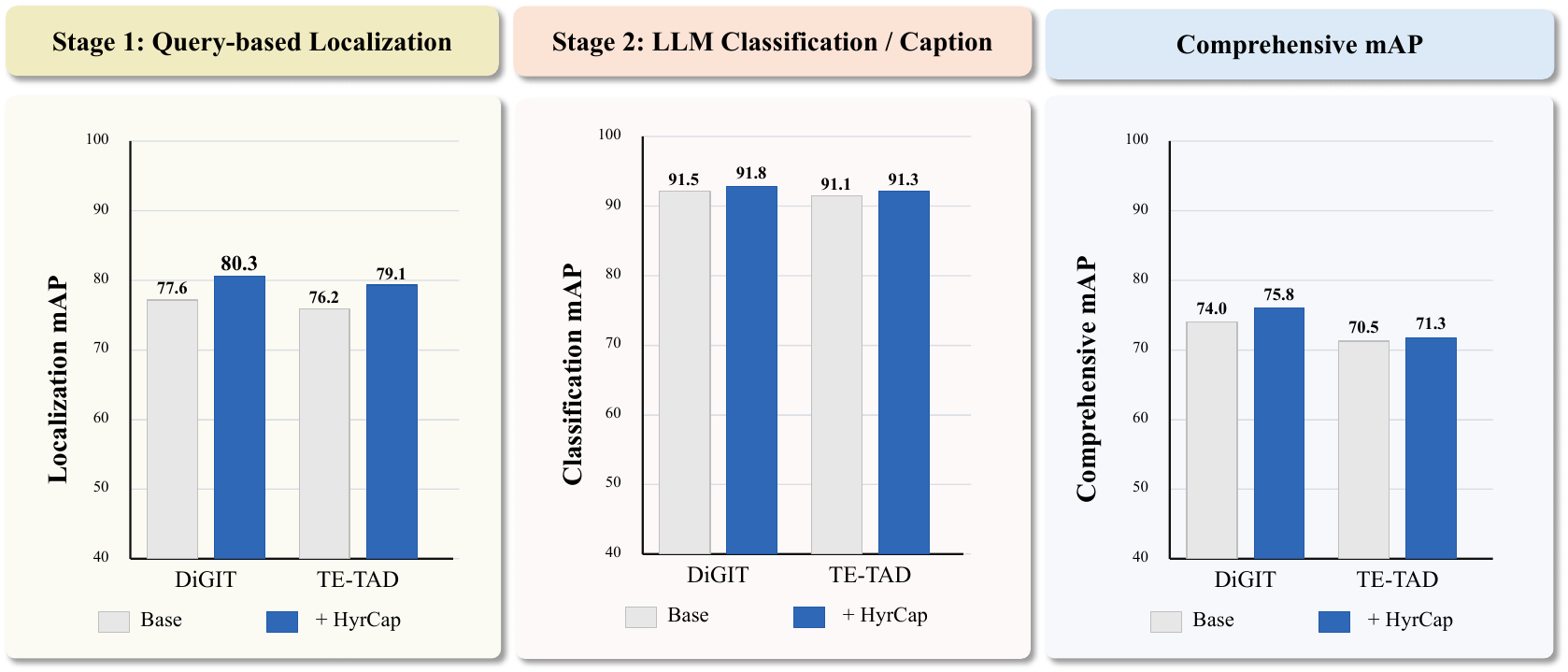}
    \caption{
    Two-Stage Evaluation.
    }
    \label{fig:stage}

\end{figure}

As shown in Fig.~\ref{fig:stage}, HyrCap consistently improves localization mAP on different query-based detectors.
For DiGIT and TE-TAD, the localization mAP increases from 77.6 and 76.2 to 80.3, 79.1, respectively.
Since the semantic model operates on localized event segments, better localization provides more reliable inputs for segment-level semantic prediction.
Under joint event evaluation, HyrCap also improves final event understanding performance, showing that calibrated proposals benefit both closed-set TAD and LLM-based semantic understanding.




\subsection{Qualitative Results}

\begin{figure}
    \centering
    \includegraphics[width=1.0\linewidth]{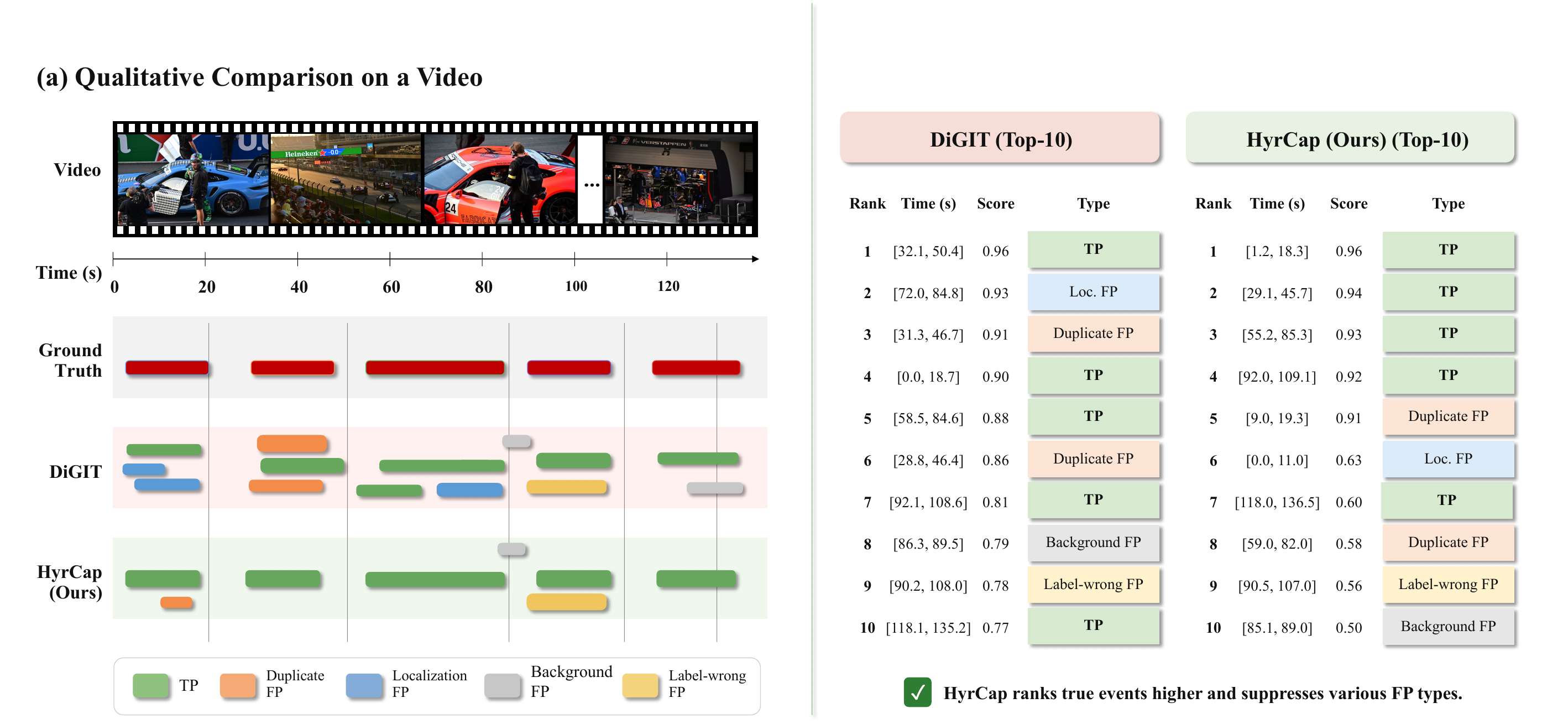}
    \caption{
    Qualitative visualization of proposal quality and ranking behavior.
    }
    \label{fig:qual}

\end{figure}

We conduct a qualitative comparison between HyrCap and the baseline DiGIT detector on representative videos. As shown in Figure \ref{fig:qual}, HyrCap generates proposals that more accurately cover the ground-truth action segments, while simultaneously reducing duplicate, background, and localization-error false positives. In contrast, DiGIT produces more redundant or misaligned proposals.

On the top-ranked proposals (Figure \ref{fig:qual}), HyrCap consistently ranks true positives higher and suppresses various types of false positives, demonstrating that its cross-representation consensus and proposal calibration effectively improve both proposal quality and ranking reliability. This visual inspection highlights HyrCap’s ability to produce temporally precise and stable event segments suitable for downstream semantic understanding.

\section{Conclusion}
In this paper, we proposed HyrCap, a proposal-centric hybrid rank-calibration framework for query-based temporal action detection and temporal event understanding. HyrCap treats dense query predictions as an over-complete temporal event hypothesis space and improves proposal quality through cross-representation consensus calibration, localization quality estimation, Proposal Redundancy Suppression, and temporal boundary refinement. Experiments on public TAD benchmarks show that HyrCap consistently improves strong query-based baselines, while ablation and qualitative results verify each component. Beyond closed-set detection, HyrCap further supports a staged semantic understanding protocol, where calibrated proposals serve as reliable event-level inputs for LLM-based classification, captioning, and reasoning. These results demonstrate the importance of high-quality temporal proposals for connecting TAD with higher-level video understanding.

\paragraph{Limitation and future work}
Although HyrCap consistently improves query-based TAD baselines, it still has several limitations. 
First, HyrCap is a proposal calibration framework and therefore depends on the quality of the base detector. 
If the base detector fails to generate candidates that cover true action instances, the missing events cannot be recovered by subsequent rank calibration or boundary refinement. 
Second, the cross-representation consensus module relies on auxiliary video representations to provide complementary temporal evidence. 
While this design improves proposal reliability, it may introduce additional feature extraction, storage, and inference cost. 
Reducing the dependence on auxiliary representations or learning more efficient consensus cues is an important direction for future work.

Third, the staged semantic understanding protocol depends on the selected event proposals, the trained projector, and the semantic model. 
Errors in temporal localization or visual-semantic projection may affect segment-level classification, captioning, and reasoning. 
Moreover, our current semantic evaluation is mainly built on predefined action categories and public TAD benchmarks, which does not fully cover open-vocabulary event discovery or complex long-form video reasoning. 
In the future, we will explore more adaptive proposal selection, more efficient proposal calibration, and stronger integration with open-vocabulary semantic models for temporal event understanding in more diverse real-world videos.


\clearpage
\appendix
\section{Implementation Details}

\subsection{Dataset Characteristics}

We evaluate HyrCap on THUMOS14~\cite{idrees2017thumos}, ActivityNet v1.3~\cite{caba2015activitynet}, and HACS-Segment~\cite{zhao2019hacs}, following the standard data splits used in prior temporal action detection works. These benchmarks provide complementary evaluation settings in terms of dataset scale, action duration, category coverage, and instance density.

THUMOS14 contains relatively short and densely occurring actions in untrimmed videos. Repeated actions and substantial background make precise boundary localization, proposal ranking, and redundancy suppression challenging. ActivityNet v1.3 covers a broader category space, with actions generally longer and more sparsely distributed than those in THUMOS14, emphasizing semantic diversity and long-duration action modeling. HACS-Segment has the largest video count among the three benchmarks, supporting evaluation on larger-scale and more diverse video distributions.

Together, these datasets evaluate the effectiveness of proposal calibration across different temporal structures, category spaces, and dataset scales. Their statistics are summarized in Table~\ref{tab:appendix_dataset_statistics}.

\begin{table*}[t]
    \centering
    \caption{Benchmark statistics.}
    \label{tab:appendix_dataset_statistics}

    \normalsize
    \renewcommand{\arraystretch}{1.35}
    \setlength{\tabcolsep}{9pt}

    \begin{tabularx}{\textwidth}{
        @{}
        l
        c
        >{\raggedright\arraybackslash}X
        >{\raggedright\arraybackslash}X
        @{}
    }
        \toprule
        \textbf{Dataset}
        & \textbf{Classes}
        & \textbf{Video splits}
        & \textbf{Action instances} \\
        \midrule

        THUMOS14
        & 20
        & \shortstack[l]{200 validation\\213 test}
        & \shortstack[l]{3,007 validation\\3,358 test} \\

        \addlinespace[4pt]

        ActivityNet v1.3
        & 200
        & \shortstack[l]{10,024 training\\4,926 validation}
        & \shortstack[l]{15,410 training\\7,654 validation} \\

        \addlinespace[4pt]

        HACS-Segment
        & 200
        & \shortstack[l]{37,613 training\\5,981 validation}
        & \shortstack[l]{105,868 training\\16,436 validation} \\

        \bottomrule
    \end{tabularx}
\end{table*}

\subsection{Training Configuration}

The base detectors are trained or evaluated following their official configurations, and HyrCap is applied to their generated proposal predictions.

\begin{sloppypar}
During proposal calibration training, we freeze the video encoders and the base detectors, and train the Cross-Representation Consensus Calibration (CPCC), Localization Quality Estimation (LQE), and Proposal Redundancy Suppression (PRS) modules. For semantic projector training, we freeze the video encoders, the base detectors, and Qwen3-14B, and train only the projector.
\end{sloppypar}

Training is conducted on four NVIDIA A100-SXM4-80GB GPUs, each with 80 GB of memory. The calibration modules are trained for 50 epochs using the AdamW optimizer with a batch size of 4. The NMS threshold is set to 0.70, and Top-K is set to 900.

\subsection{Measurement Settings}

We use DiGIT as the baseline to measure the computational requirements of HyrCap. Measurements are performed on a single NVIDIA RTX 4090 GPU using pre-extracted video features, FP32 precision, and a batch size of one. Inference time for the detection and calibration stages is averaged over all THUMOS14 test videos, whose mean duration is 217.27 seconds.

\section{Robustness Analysis}

\subsection{Temporal-Structure Perturbations}

To evaluate robustness to temporal-structure changes, we construct a temporally perturbed evaluation set by moving a segment from the latter part of each video to the beginning. This operation changes the ordering of the video content and its temporal context. We evaluate HyrCap on the clean and perturbed sets using the THUMOS14 detection metrics. Average mAP is computed over five tIoU thresholds from 0.3 to 0.7 with a step of 0.1.

As shown in Table~\ref{tab:appendix_temporal_perturbations}, HyrCap achieves an average mAP of 73.3 on the perturbed set, a decrease of 2.3 points from the clean set. AP decreases by 2.2, 2.1, and 1.5 points at tIoU thresholds of 0.3, 0.5, and 0.7, respectively. These results show that HyrCap maintains stable detection performance under temporal rearrangement.

\begin{table*}[t]
    \centering
    \caption{Performance under temporal-structure perturbations
    on THUMOS14.}
    \label{tab:appendix_temporal_perturbations}

    \normalsize
    \renewcommand{\arraystretch}{1.35}
    \setlength{\tabcolsep}{9pt}

    \begin{tabular*}{\textwidth}{
        @{\extracolsep{\fill}}lcccc@{}
    }
        \toprule
        \textbf{Evaluation setting}
        & \textbf{AP@0.3}
        & \textbf{AP@0.5}
        & \textbf{AP@0.7}
        & \textbf{Avg. mAP} \\
        \midrule

        Clean set
        & 88.1 & 79.1 & 54.6 & 75.6 \\

        \addlinespace[3pt]

        Temporally perturbed set
        & 85.9 & 77.0 & 53.1 & 73.3 \\

        \addlinespace[3pt]

        Change
        & $-2.2$ & $-2.1$ & $-1.5$ & $-2.3$ \\

        \bottomrule
    \end{tabular*}
\end{table*}

\subsection{Proposal-Level Perturbations}

We further evaluate robustness to imperfect proposal inputs using two perturbations: $\pm 5\%$ boundary jitter and random removal of $5\%$ of proposals. Boundary jitter introduces temporal localization noise, while proposal removal reduces the available candidate evidence.

As shown in Table~\ref{tab:appendix_proposal_perturbations}, HyrCap achieves average mAP values of 73.9 and 74.0 under boundary jitter and proposal removal, respectively, corresponding to decreases of 1.7 and 1.6 points. These results demonstrate robustness to proposal-boundary noise and missing proposals.

\begin{table*}[t]
    \centering
    \caption{Performance under proposal-level perturbations
    on THUMOS14.}
    \label{tab:appendix_proposal_perturbations}

    \normalsize
    \renewcommand{\arraystretch}{1.35}
    \setlength{\tabcolsep}{9pt}

    \begin{tabular*}{\textwidth}{
        @{\extracolsep{\fill}}lcc@{}
    }
        \toprule
        \textbf{Proposal setting}
        & \textbf{Avg. mAP}
        & \textbf{Change} \\
        \midrule

        Clean proposals
        & 75.6 & --- \\

        \addlinespace[3pt]

        $\pm 5\%$ boundary jitter
        & 73.9 & $-1.7$ \\

        \addlinespace[3pt]

        Random removal of $5\%$ of proposals
        & 74.0 & $-1.6$ \\

        \bottomrule
    \end{tabular*}
\end{table*}

\clearpage
\section*{NeurIPS Paper Checklist}

\begin{enumerate}

\item {\bf Claims}
    \item[] Question: Do the main claims made in the abstract and introduction accurately reflect the paper's contributions and scope?
    \item[] Answer: \answerYes{}
    \item[] Justification: The abstract and introduction state the main contributions of HyrCap, including hybrid rank calibration, proposal refinement, and staged semantic understanding. These claims are supported by the method design, ablation studies, standard TAD evaluation, and staged semantic evaluation.

\item {\bf Limitations}
    \item[] Question: Does the paper discuss the limitations of the work performed by the authors?
    \item[] Answer: \answerYes{}
    \item[] Justification: The paper discusses the scope and limitations of HyrCap, including its reliance on pretrained video features, base query-based detectors, and the quality of generated proposals for downstream semantic understanding.

\item {\bf Theory assumptions and proofs}
    \item[] Question: For each theoretical result, does the paper provide the full set of assumptions and a complete (and correct) proof?
    \item[] Answer: \answerNA{}
    \item[] Justification: The paper does not present theoretical theorems or formal proofs. The equations are used to define the proposed empirical framework and evaluation protocol.

\item {\bf Experimental result reproducibility}
    \item[] Question: Does the paper fully disclose all the information needed to reproduce the main experimental results of the paper to the extent that it affects the main claims and/or conclusions of the paper (regardless of whether the code and data are provided or not)?
    \item[] Answer: \answerYes{}
    \item[] Justification: The paper specifies the datasets, feature representations, base detectors, evaluation metrics, module designs, training settings, and post-processing procedures needed to reproduce the main results. Additional implementation details and configurations are provided in the experimental section and supplemental material.

\item {\bf Open access to data and code}
    \item[] Question: Does the paper provide open access to the data and code, with sufficient instructions to faithfully reproduce the main experimental results, as described in supplemental material?
    \item[] Answer: \answerYes{}
    \item[] Justification: The experiments are conducted on public datasets and public feature/model resources. We will release anonymized code, configuration files, and reproduction scripts for the proposed modules and evaluation protocols.

\item {\bf Experimental setting/details}
    \item[] Question: Does the paper specify all the training and test details (e.g., data splits, hyperparameters, how they were chosen, type of optimizer) necessary to understand the results?
    \item[] Answer: \answerYes{}
    \item[] Justification: The paper reports the datasets, splits, feature representations, base detectors, evaluation metrics, module settings, and ablation configurations used in the experiments. Training and inference details are included in the implementation details and supplemental material.

\item {\bf Experiment statistical significance}
    \item[] Question: Does the paper report error bars suitably and correctly defined or other appropriate information about the statistical significance of the experiments?
    \item[] Answer: \answerNo{}
    \item[] Justification: The paper reports standard TAD benchmark results following common practice, but does not include repeated-run error bars because training and evaluating multiple large video detection models is computationally expensive. The main claims are supported by consistent improvements across datasets, base detectors, and ablation settings.

\item {\bf Experiments compute resources}
    \item[] Question: For each experiment, does the paper provide sufficient information on the computer resources (type of compute workers, memory, time of execution) needed to reproduce the experiments?
    \item[] Answer: \answerYes{}
    \item[] Justification: The paper provides the main compute settings used for training and evaluation, including GPU-based training, pretrained feature extraction, and lightweight module training. Additional resource details are included in the implementation section or supplemental material.
    
\item {\bf Code of ethics}
    \item[] Question: Does the research conducted in the paper conform, in every respect, with the NeurIPS Code of Ethics \url{https://neurips.cc/public/EthicsGuidelines}?
    \item[] Answer: \answerYes{}
    \item[] Justification: The work uses public video understanding benchmarks and publicly available pretrained models or features. The research is conducted for temporal action detection and event understanding, and does not involve private data collection or human-subject experiments.

\item {\bf Broader impacts}
    \item[] Question: Does the paper discuss both potential positive societal impacts and negative societal impacts of the work performed?
    \item[] Answer: \answerYes{}
    \item[] Justification: The paper discusses potential benefits for video retrieval, event annotation, and assistive video understanding. It also acknowledges that improved video understanding systems may be misused in surveillance or privacy-sensitive applications.
    
\item {\bf Safeguards}
    \item[] Question: Does the paper describe safeguards that have been put in place for responsible release of data or models that have a high risk for misuse (e.g., pre-trained language models, image generators, or scraped datasets)?
    \item[] Answer: \answerNA{}
    \item[] Justification: The paper does not introduce a new high-risk pretrained foundation model, image generator, or scraped dataset. It uses existing public datasets and pretrained resources for research evaluation.

\item {\bf Licenses for existing assets}
    \item[] Question: Are the creators or original owners of assets (e.g., code, data, models), used in the paper, properly credited and are the license and terms of use explicitly mentioned and properly respected?
    \item[] Answer: \answerYes{}
    \item[] Justification: The paper cites the original works for the datasets, pretrained video features, base detectors, and language model resources used in the experiments. We follow the corresponding licenses and terms of use for public assets such as THUMOS14, ActivityNet v1.3, HACS-Segment, I3D, InternVideo2, V-JEPA2, DiGIT, TE-TAD, TadTR, and Qwen3-14B.

\item {\bf New assets}
    \item[] Question: Are new assets introduced in the paper well documented and is the documentation provided alongside the assets?
    \item[] Answer: \answerYes{}
    \item[] Justification: The paper introduces code, configuration files, calibration scripts, and evaluation scripts for HyrCap rather than a new dataset. These assets will be documented with usage instructions, environment requirements, and reproduction commands.

\item {\bf Crowdsourcing and research with human subjects}
    \item[] Question: For crowdsourcing experiments and research with human subjects, does the paper include the full text of instructions given to participants and screenshots, if applicable, as well as details about compensation (if any)? 
    \item[] Answer: \answerNA{}
    \item[] Justification: The paper does not involve crowdsourcing experiments or research with human subjects.

\item {\bf Institutional review board (IRB) approvals or equivalent for research with human subjects}
    \item[] Question: Does the paper describe potential risks incurred by study participants, whether such risks were disclosed to the subjects, and whether Institutional Review Board (IRB) approvals (or an equivalent approval/review based on the requirements of your country or institution) were obtained?
    \item[] Answer: \answerNA{}
    \item[] Justification: The paper does not involve human-subject experiments, user studies, or newly collected human data, so IRB approval is not applicable.

\item {\bf Declaration of LLM usage}
    \item[] Question: Does the paper describe the usage of LLMs if it is an important, original, or non-standard component of the core methods in this research? Note that if the LLM is used only for writing, editing, or formatting purposes and does \emph{not} impact the core methodology, scientific rigor, or originality of the research, declaration is not required.
    \item[] Answer: \answerYes{}
    \item[] Justification: The paper explicitly describes the use of Qwen3-14B in the staged semantic understanding protocol. The LLM is used for segment-level semantic prediction, captioning, or reasoning after proposal localization, and its confidence is not used for proposal ranking.

\end{enumerate}

\end{document}